\pdfoutput=1
\documentclass{article}

\usepackage{arxiv}

\usepackage[utf8]{inputenc} 
\usepackage[T1]{fontenc}    
\usepackage{hyperref}       
\usepackage{url}            
\usepackage{booktabs}       
\usepackage{lipsum}		
\usepackage{graphicx}
\usepackage{natbib}
\usepackage{doi}
\usepackage{amsmath,amssymb,amsfonts,bm}
\usepackage{amsmath}
\usepackage{array}

\usepackage{fixltx2e}

\usepackage{stfloats}

\title{Learning User Interaction Forces using Vision for a Soft Finger Exosuit}

\author{
Mohamed Irfan Refai$^{1,2,*}$,
Abdulaziz Y. Alkayas$^{2,3}$,
Anup Teejo Mathew$^3$, \\
 \textbf{Federico Renda$^3$, Thomas George Thuruthel$^{2}$} \\
        Email: \texttt{$\{$*m.i.mohamedrefai$\}$}@utwente.nl \\
$^1$Department of Biomechanical Engineering, University of Twente, The Netherlands, \\
$^2$Department of Computer Science, University College London, The UK, \\  $^3$Khalifa University, UAE.}

\begin{document}
\maketitle

\begin{abstract}
Wearable assistive devices are increasingly becoming softer. Modelling their interface with human tissue is necessary to capture transmission of dynamic assistance. However, their nonlinear and compliant nature makes both physical modeling and embedded sensing challenging. In this paper, we develop a image-based, learning-based framework to estimate distributed contact forces for a finger-exosuit system. We used the SoRoSim toolbox to generate a diverse dataset of exosuit geometries and actuation scenarios for training. The method accurately estimated interaction forces across multiple contact locations from low-resolution grayscale images, was able to generalize to unseen shapes and actuation levels, and remained robust under visual noise and contrast variations. We integrated the model into a feedback controller, and found that the vision-based estimator functions as a surrogate force sensor for closed-loop control. This approach could be used as a non-intrusive alternative for real-time force estimation for exosuits.
\end{abstract}



\section{Introduction}
\label{sec:introduction}
Activities of the upper extremity such as reaching and grasping offers individuals independence in interacting with the environment in daily life. However, these activities may be impaired due to aging or trauma \citep{Holt2013, Saes2022}. Wearable assistive devices or exosuits can offer continuous support as needed for the individual in daily life \citep{Bardi2022}. Exosuits made of rigid structures allow better interfacing with the fingers and hand, resulting in efficient transmission of assistance to the user \citep{Bardi2022, Scherb2023}. However, such exosuits are generally bulky and have a large form factor. Moreover, any misalignment with the finger or wrist joints can result in resistance or inertial effects \citep{AsbeckExos2014}. These factors result in user discomfort over time. 

Exosuits made from softer materials can be rather lightweight and conform better to the biological joints \citep{AsbeckExos2014}. Therefore, they can be more comfortable than rigid exosuits. However, modelling softer exosuits (or robots) can be quite challenging due to their complex dynamics and continuous interactions with the environment \citep{Rus2015}. Modelling soft exosuit dynamics and interaction forces with the user is necessary to improve efficiency of these systems during long term use \citep{Massardi2022}. 


Measuring the forces between an exosuit and the user is generally challenging. Studies have utilized embedded sensing to measure interactions between exosuit and user or a soft robot with its environment. Pressure mats or force sensitive resistors have been used to extrapolate the interaction between the user and the assistive device \citep{Andrikopoulos2015, Xiloyannis2018}. Additionally, embedded solutions such as fibre Bragg grating (FBG) sensors \citep{khan2017force, bai2020stretchable}, optical transducers \citep{noh2016multi}, and pressure sensors \citep{haraguchi2015pneumatically} have been previously used for force estimation on soft bodies. However, such methods are highly limited in terms of sensing location, can impede the motion of the soft body, and are not applicable to all design forms.

\begin{figure}[t!]
    \centering
    \includegraphics[trim = 0 5cm 0 0,clip,width=0.8\textwidth]{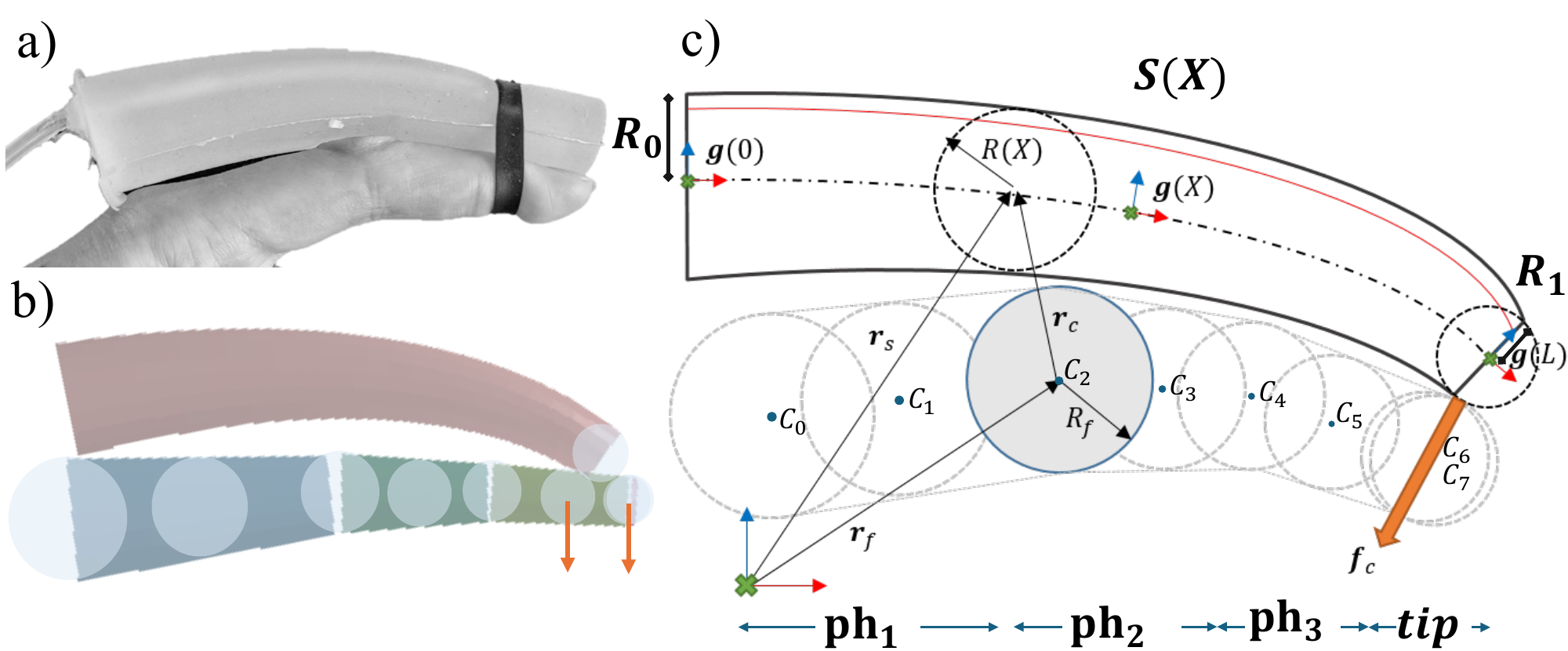}
    \caption{ (a) Modelling interaction forces of a soft exosuit using vision-based learning techniques validated using a (b) simulated model. (c) The exosuit is modelled as a Cosserat rod, and the finger as three rigid links connected via rotational joints (RRR). The red line on top of the exosuit shows the line of actuation. Spheres were defined along the exosuit and finger ($C_i$) to approximate the contact mechanics.}
    \label{fig_use_case}
\end{figure}

Alternatively, accurate forward models of soft-bodied systems can be used to estimate interaction forces when combined with optimization techniques. Model-based approaches have been used to estimate external forces from observed states or deformations in soft robotic systems that deform with constant curvature \citep{camarillo2008mechanics,gao2016mechanical,6094526,bajo2016hybrid}. Advanced modelling appraoches are however required for state estimation of complex soft-bodied systems without embedded sensors. Although, Finite Element Methods (FEM) can provide a comprehensive analysis of the interaction dynamics between the exosuit and the user \citep{Faure2012, Menager2025}, they can be computationally intensive to simulate and optimize. Simplified (In-)Finite Dimensional Models of soft robots offer efficient modelling alternative for interaction and control analyses \citep{Santina2023, Armanini2023}. The SoRoSim toolbox \citep{Mathew2023} implements such an approach to model multibody hybrid structures using the Geometric Variable Strain (GVS) model \citep{Boyer_TRO2020, renda2020geometric, Mathew2024}. 

Nonetheless, model-based techniques still require identification of system parameters from real-world data which is challenging for complex nonlinear behaviour of viscoelastic materials. Learning-based techniques have been shown to be successful in abstracting soft-robotics dynamics from images \citep{Monteiro2023}. In this simulation study, we test whether similar techniques can be applied to our problem. Therefore, we develop a data-driven modelling framework for estimating the interaction forces between a soft exosuit and the finger it is assisting. Our modelling and perception approach relies only on visual data that can be obtained using simple external cameras. The framework is an extension of studies which estimate tip forces \citep{chen2024vision} and soft-bodied deformation \citep{Huang2024} using convolutional neural networks. Unlike these studies, we show how such data-driven methods can generalize to new shapes and estimate contact forces at multiple locations without any explicit optimization techniques. We validate our model on a finger-exosuit system modelled using the SoRoSim toolbox. The exosuit, composed of a soft, actuated body, facilitates bending of the human finger when actuated. We test the robustness of the technique to varied visual noise and perform further validation on closed-loop control tasks.

\section{Methods}
\label{sec:methods}
Section \ref{sec_finger_exosuit} describes the finger-exosuit design. The CNN force estimator architecture is described in Section \ref{sec_CNN_archi}, followed by the synthesis of datasets for learning in Section  \ref{sec_data_synthesis}. We assess the generalizability of the CNN force estimator and use in a feedback controller in Sections \ref{sec_generalizability} and \ref{sec_feedback_controller} respectively. Finally, we describe the results presented in Section \ref{sec_analysis_results}.

\begin{table}[t!]
	\caption{Initial design parameters of the finger-exosuit}
	\centering
	\begin{tabular}{|l|c|c|c|c|}
		\hline
		\textbf{Item} & 
		\begin{tabular}[c]{@{}c@{}}\textbf{Length} \\ (cm) \end{tabular}  & \begin{tabular}[c]{@{}c@{}}\textbf{Radius Base} \\ (cm) \end{tabular} & \begin{tabular}[c]{@{}c@{}}\textbf{Radius Tip} \\ (cm) \end{tabular} & \begin{tabular}[c]{@{}c@{}}\textbf{Joint } \\ \textbf{Stiffness} (Nm/rad) \end{tabular}     \\ \hline
		exosuit    & 10   & 0.58  & 0.4   & --   \\ \hline
		$ph_1$     & 4.5  & 1.08  & 0.695 & 0.04 \\ \hline
		$ph_2$     & 2.53 & 0.695 & 0.51  & 0.03 \\ \hline
		$ph_3$     & 2.36 & 0.51  & 0.435 & 0.02 \\ \hline
		tip        & 0.1  & 0.435 & 0.4   & -- \\ \hline		
	\end{tabular}
	\label{tab_system_dim}
\end{table}

\subsection{Modelling the finger-exosuit system} \label{sec_finger_exosuit}
\begin{figure*}[t!]
    \centering
    \includegraphics[trim = 0 8cm 0 0,clip,scale = 0.5]{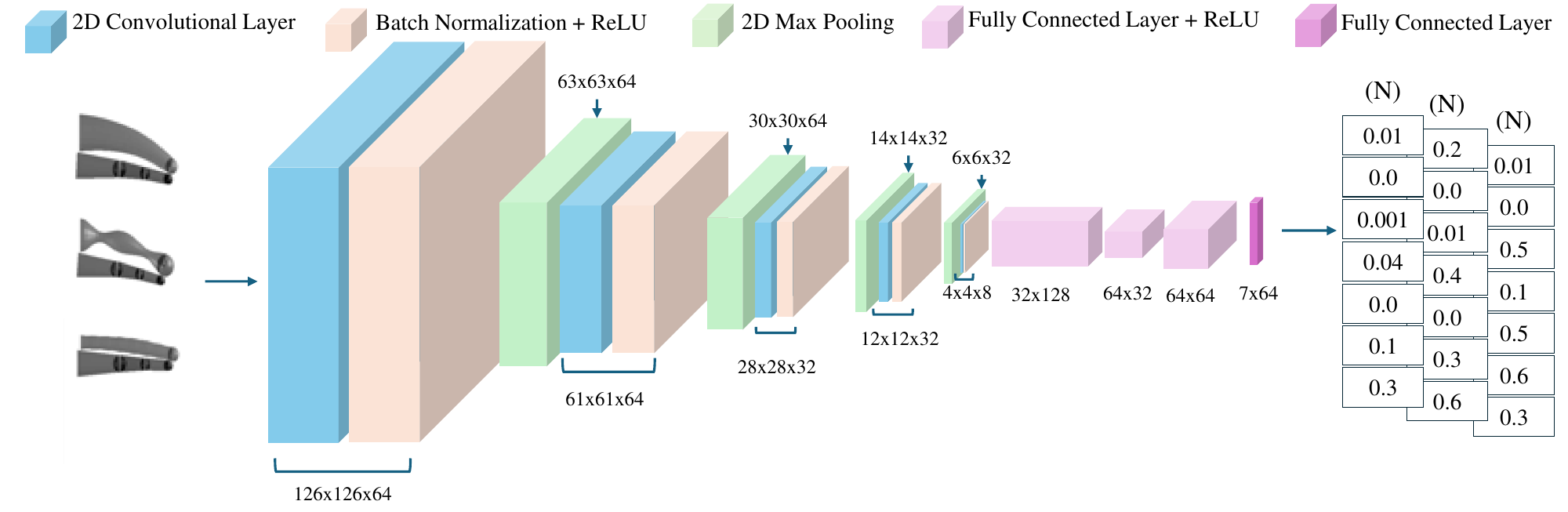}
    \caption{Architecture of the CNN mapping low resolution (128x128) grayscale images of the finger-exosuit sytem to 1x7 contact forces. The contact forces are measured on $C_i$, where $i \in \{1, \; 2,\; ..., \; 7\}$. }
    \label{fig_architecture}
\end{figure*}
Figure \ref{fig_use_case}a depicts the exosuit worn in parallel with the index finger. The finger-exosuit system was modelled using SoRoSim toolbox\footnote{\url{https://github.com/SoRoSim/SoRoSim.git}} \citep{Mathew2023}.  
\subsubsection{Kinematics of the exosuit}
The exosuit was designed to simulate a soft body made with silicone, and can be actuated along a line of actuation placed on the surface opposite to the finger. The soft exosuit is modeled as a Cosserat rod, which is a continuum of rigid cross-sections along a curvilinear coordinate $X \in [0, L]$, with $L$ the total length. Each cross-section carries an orthonormal frame, yielding the rod’s configuration as a directed spatial curve $\boldsymbol{g}(X) \in SE(3)$, represented by a homogeneous transformation matrix:

\begin{equation}
    \boldsymbol{g}(X) = \left[ \begin{array}{cc} \boldsymbol{R} & \boldsymbol{r} \\ \boldsymbol{0}^{1\times 3} & 1 \end{array} \right] 
    \label{eqn::homogenousTM_g}
\end{equation}
\noindent where $\boldsymbol{r}(X) \in \mathbb{R}^3$ is the position of the moving frame’s origin, and $\boldsymbol{R}(X) \in SO(3)$ defines its orientation relative to the spatial frame, with the local x-axis orthogonal to the cross-sectional plane. Differentiating equation (\ref{eqn::homogenousTM_g}) with respect to space $(\cdot)'$ and time $\dot{(\cdot)}$ gives:
\begin{equation}\label{eqn::gprime} \boldsymbol{g}'(X)=\boldsymbol{g}\hat{\boldsymbol{\xi}}; \hspace{0.7cm} \dot{\boldsymbol{g}}(X)=\boldsymbol{g}\hat{\boldsymbol{\eta}}
\end{equation}
\noindent The Cosserat rod’s strain is denoted by $\hat{\boldsymbol{\xi}}(X)$, and its velocity twist by $\hat{\boldsymbol{\eta}}(X)$, capturing translational and rotational velocities along the rod. The operator $\hat{(\bullet)}$ maps $\mathbb{R}^6$ to $\mathfrak{se}(3)$. Equating mixed partial derivatives in space $X$ and time $t$ reveals the following relationship:
\begin{equation}\label{eqn::vel}   \boldsymbol{\eta}'=\dot{\boldsymbol{\xi}}-\text{ad}_{\boldsymbol{\xi}}\boldsymbol{\eta}
\end{equation}
\noindent where \(\text{ad}_{\boldsymbol{\xi}}\) is the adjoint operator of \(\boldsymbol{\xi}\). 

The pose model is obtained by integrating eqn. (\ref{eqn::gprime}), whereas the velocity model is obtained by integrating eqn. (\ref{eqn::vel}):
\begin{equation}\label{eq::int_kinem_pos}
\bm{g}(X) = \text{exp} \left( \widehat{\bm{\Omega}}(X) \right) \; \text{,}
\end{equation}
\begin{equation}\label{eq::int_kinem_eta}
    \boldsymbol{\eta}(X) =  \text{Ad}_{\boldsymbol{g}^{-1}}\int_0^{X} \text{Ad}_{\boldsymbol{g}} \dot{\boldsymbol{\xi}} ds
\end{equation}
The Adjoint of the homogeneous matrix $\boldsymbol{g}$ is denoted $\text{Ad}_{\boldsymbol{g}}$, and the Magnus expansion of $\boldsymbol{\xi}(X)$ by $\boldsymbol{\Omega}$. The generalized coordinates are introduced, and the continuous strain field is discretized using a finite set of strain bases:

\begin{equation}
    \boldsymbol{\xi}(X) = \boldsymbol{\Phi}_\xi(X)\boldsymbol{q} + \boldsymbol{\xi}^*(X)
    \label{eqn::discretizedStrain}
\end{equation}
Here, $\boldsymbol{\Phi}_\xi(X) \in \mathbb{R}^{6 \times n}$ defines the strain basis, $\boldsymbol{q} \in \mathbb{R}^n$ is the vector of generalized coordinates, and $\boldsymbol{\xi}^*(X)$ denotes the natural (unstressed) strain. Substituting equation (\ref{eqn::discretizedStrain}) into (\ref{eqn::vel}) yields the velocity model:
\begin{equation}
    \boldsymbol{\eta}(X) = \mathrm{Ad}_{\bm{g}(X)}^{-1} \int_0^X \mathrm{Ad}_{\bm{g}} \bm{\Phi}_{\xi} ds \dot{\bm{q}} = \boldsymbol{J}(\boldsymbol{q},X)\dot{\boldsymbol{q}}
\end{equation}
Here, the geometric Jacobian is represented by \(\boldsymbol{J} \in \mathbb{R}^{6 \times n}\). Despite being analytical, eqns. (\ref{eq::int_kinem_pos}) and (\ref{eq::int_kinem_eta}) cannot be directly calculated. \citep{Mathew2024,GeomExct_Renda2022} offers a quadrature approximation of the Magnus expansion \(\bm{\Omega}\) in a recursive formulation of the kinematic equations. 
The exosuit uses a nodal strain basis akin to the quadratic FEM basis, with three quadrature points \citep{Mathew2024} allowing degrees of freedom including extension along the local x-axis and bending about the y-axis.

\subsubsection{Kinematics of the finger}  
The finger was modelled as a rigid passive three-bar linkage with rotational joints (RRR) as seen in Table \ref{tab_system_dim}. The length of the links ($ph_1$, $ph_2$, $ph_3$) were 4.5, 2.5, and 2.4~cm respectively, and their base radii were 1.1, 0.7, and 0.5~cm respectively \citep{Kuo2012fingerdim}. The stiffness of the joint between the base and $ph_1$ was 0.04 Nm/rad, and 0.03, and 0.02 Nm/rad respectively for the successive joints \citep{Shi2020stiff}.
The linkage modeling follows classical rigid robot modeling techniques \citep{Lynch_Park_2017}, with external loads defined by contact forces included.  
Additional contact points at the fingertip were modeled using a fixed link of negligible length.

The full system comprises 25 degrees of freedom: 22 for the exosuit’s kinematics and 3 for the finger joint rotations. The rotation of the first finger joint was restricted to the range of -10$^{\circ}$ to 90$^{\circ}$, while the second and third joints were constrained between -5$^{\circ}$ and 90$^{\circ}$.

\subsubsection{Contact forces} \label{sec_contact_forces}

To simulate exosuit–finger interaction, a simplified interference contact model was used, approximating contact as between two spheres (see Fig. \ref{fig_use_case}). The model is defined as follows:

\begin{equation}\label{eqn::ContactModel}
    \bm{f}_{C_i} = \begin{cases}
    kd_i \dfrac{\bm{r}_{c_i}}{\lVert \bm{r}_{c_i} \rVert} &\text{if} \hspace{4pt} d_i > 0 \\
    0 &\text{if} \hspace{4pt} d_i < 0
    \end{cases},
\end{equation}
where \(\bm{r}_{c_i} = \bm{r}_{s}(X) - \bm{r}_{f_i}\), \(d_i = R_{f_i} + R(X) - \lVert \bm{r}_{c_i} \rVert\), and \(k\) represents the contact stiffness of the model. The contact force model includes only a normal component, with tangential friction neglected for simplicity. The SoRoSim toolbox identified a total of 42 gaussian quadrature points along the exosuit, each of which had a contact sphere. For the finger, two contact spheres were defined for each rigid link; one at the base and another at the Centre of Mass, resulting in eight contact spheres ($C_i$, where $i \in \{0, \; 1, \; ..., \; 7\}$). We found this discretization to be suitable with the limited motion of the rigid bodies. The system's static equilibrium was determined by solving for the contact interactions between the finger and the exosuit.

\subsubsection{Statics} \label{sec_statics}

The generalized static equation is obtained by projecting the static equilibrium equations of the soft exosuit and the rigid finger onto the space of generalized coordinates using the geometric Jacobian, following D'Alembert's principle \citep{renda2020geometric}:

\begin{equation}\label{eqn::genStcEqn}
     \boldsymbol{K}\boldsymbol{q} = \boldsymbol{B}(\boldsymbol{q})u + \boldsymbol{F}(\boldsymbol{q}).
\end{equation}

\noindent Here, $\boldsymbol{K} \in \mathbb{R}^{n \times n}$ is the stiffness matrix, $\boldsymbol{B}(\boldsymbol{q}) \in \mathbb{R}^n$ is the actuation matrix, $u$ is the pneumatic actuation force (shown as the red line in Fig. \ref{fig_use_case}), and $\boldsymbol{F}(\boldsymbol{q}) \in \mathbb{R}^n$ represents generalized external (contact) forces. The Young’s modulus was 30~kPa, representing that of Ecoflex.

To solve the full system model, root-finding methods are applied to eqn. (\ref{eqn::genStcEqn}). The actuation force is incrementally increased, using each equilibrium solution as the initial guess for the next step, ensuring stability and computational efficiency as the load gradually increases.

\subsection{CNN Force estimator architecture} \label{sec_CNN_archi}
The CNN model (Fig. \ref{fig_architecture}) was designed to map the image of the finger-exosuit system with the interaction forces defined along the finger. We employed a single network to estimate the normal forces along the finger at seven predefined points in the simulation. The model must therefore infer the kinemato-static behavior of the soft robot (for any shape) and additionally estimate the contact locations and interaction forces. The model architecture has 5 convolutional layers with 64, 64, 32, 32, and 8 filters respectively, each with a 3x3 kernel. Each of the first four layers were followed with a batch normalization, a ReLU activation, and a 2x2 max pooling layer. Three fully connected layers (32, 32, 64 neurons respectively) were placed next, each with a ReLU activation layer. Finally, a fully connected layer with 64 neurons was mapped to the 1x8 force outputs. $C_0$ was always 0 and therefore ignored for later analyses. There were a total of 78,847 trainable parameters.   

The network utilized an Adam optimizer with an L1 loss function. The batch size was set to 64. The learning rate was initialized to $2e-4$, which was reduced by 0.1 when plateauing (after at least 10 epochs). Training was stopped when validation loss (measured every 5 iterations) did not improve after 5 instances. 

Training was done using pytorch library on an Ubuntu system running on Intel(R) Xeon(R) CPU (2.20GHz) with a Titan XP GPU (GP102, NVIDIA, 33MHz clock).
\begin{table}[t!]
	\caption{Bounds for each parameter}
	\centering
	\begin{tabular}{|l|c|c|c|c|}
		\hline
		\textbf{Parameter} & \textbf{Lower Bd.} & \textbf{Upper Bd.} & \textbf{Steps} & \textbf{Eqn.} \\ \hline
		
		\(L\)    & 7 cm   & 11 cm  & 0.01 cm   & --   \\ \hline
		$R_0, R_1$     & 0.4 cm  & 1 cm  & 0.01 cm & (\ref{eqn_R01}) \\ \hline
		$A$, $B$     & -15 & 1e-3 & 0.1  & (\ref{eqn_Rs1}), (\ref{eqn_Rcc}) \\ \hline
		$C,D$     & 1e-4 & 5  & 0.01 & (\ref{eqn_RSC}) \\ \hline
		$m_n$        & 0.001  & 1 & 5e-3   & (\ref{eqn_RLP}) \\ \hline		
	\end{tabular}
	\label{tab_parameter_bounds}
\end{table}

\subsection{Simulation of dataset for learning} \label{sec_data_synthesis}
We generated different exosuit shapes by varying the length \(L\) and the shape of the exosuit surface $S(X)$, and actuated them to simulate data for training the CNN. We formulated five different approaches to model $S(X)$ as follows:

\begin{equation}
    R_{0,1}: S(X) =  R_0 + X_L \cdot (R_1 - R_0), \label{eqn_R01}
\end{equation}
\begin{equation}
    R_{s}: S(X) =  R_0 + \frac{R_0}{20} sin(A \cdot pi \cdot X_L), \label{eqn_Rs1}
\end{equation}
\begin{equation}
    R_{cc}: S(X) =  \sum_{n=0}^{6} B_{n} cos(\pi \cdot n \cdot X_L), \label{eqn_Rcc}
\end{equation}
\begin{equation}
\begin{split}
    R_{sc}: S(X) &= R_0 + R_s (sin(C \cdot \pi \cdot X_L + \phi_C)  
    \\ &+ cos(D \cdot \pi \cdot X_L + \phi_D)), \label{eqn_RSC}
\end{split}
\end{equation}
\begin{equation}
    R_{LP}: S(X) =\sum_{n=0}^{N-1} \frac{2m_n}{3^n} P_n(X_L). \label{eqn_RLP}
\end{equation}

Eqn. (\ref{eqn_R01}) models $S(X)$ linearly from the base ($R_0$) to tip ($R_1$) of the exosuit. Here, $X_L$ is the normalized length along the axis of the exosuit. The two parameters ($R_0$, $R_1$) can be adjusted to simulate different exosuit shapes. 
In eqn. (\ref{eqn_Rs1}), the frequency factor $A$ can be varied to generate different exosuit shapes. Eqn. (\ref{eqn_Rcc}) defines the $S(X)$ as a sum of 6 cosine waves, where the amplitude ($B_n$) of each can be varied. Eqn. (\ref{eqn_RSC}) formulates $S(X)$ as the sum of a sine and a cosine wave, with four degrees of freedom (frequency ($C$, $D$) and phase ($\phi_C$, $\phi_D$)). Finally, eqn.  (\ref{eqn_RLP}) denotes $S(X)$ as a sum of Legendre Polynomials (LP), where the amplitude of the LPs ($m_n$) could be varied. To prevent very small exosuit thickness, the minimum radius of the exosuit was limited to 0.4 cm. By combining the $S(X)$ with \(L\), and varying either or both, we had 11 scenarios for simulating a unique exosuit. The bounds of each parameter is mentioned in Table \ref{tab_parameter_bounds}. We randomly selected unique combinations from these parameter bounds to generate 125 different instances for each scenario, resulting in 1375 unique exosuit shapes. For each shape, we actuated the finger-exosuit system with force of 0 to 4 N and back to 0 N. From this actuation, we interpolated 50 frames resulting in a total of 68,750 frames. For a few instances, the exosuit reached unrealistic poses after actuation. We filtered the poses where the exosuit tip was below the finger tip, resulting in a total of 65,637 frames of the finger-exosuit system for learning. The simulated frames were then randomized and split into training (60\%), validation (20\%), and testing (20\%) sets. All the results shown in the next section are obtained on the test set. 

\subsection{Generalizability} \label{sec_generalizability}
\begin{figure}[t!]
    \centering
    \includegraphics[width = 0.7\textwidth]{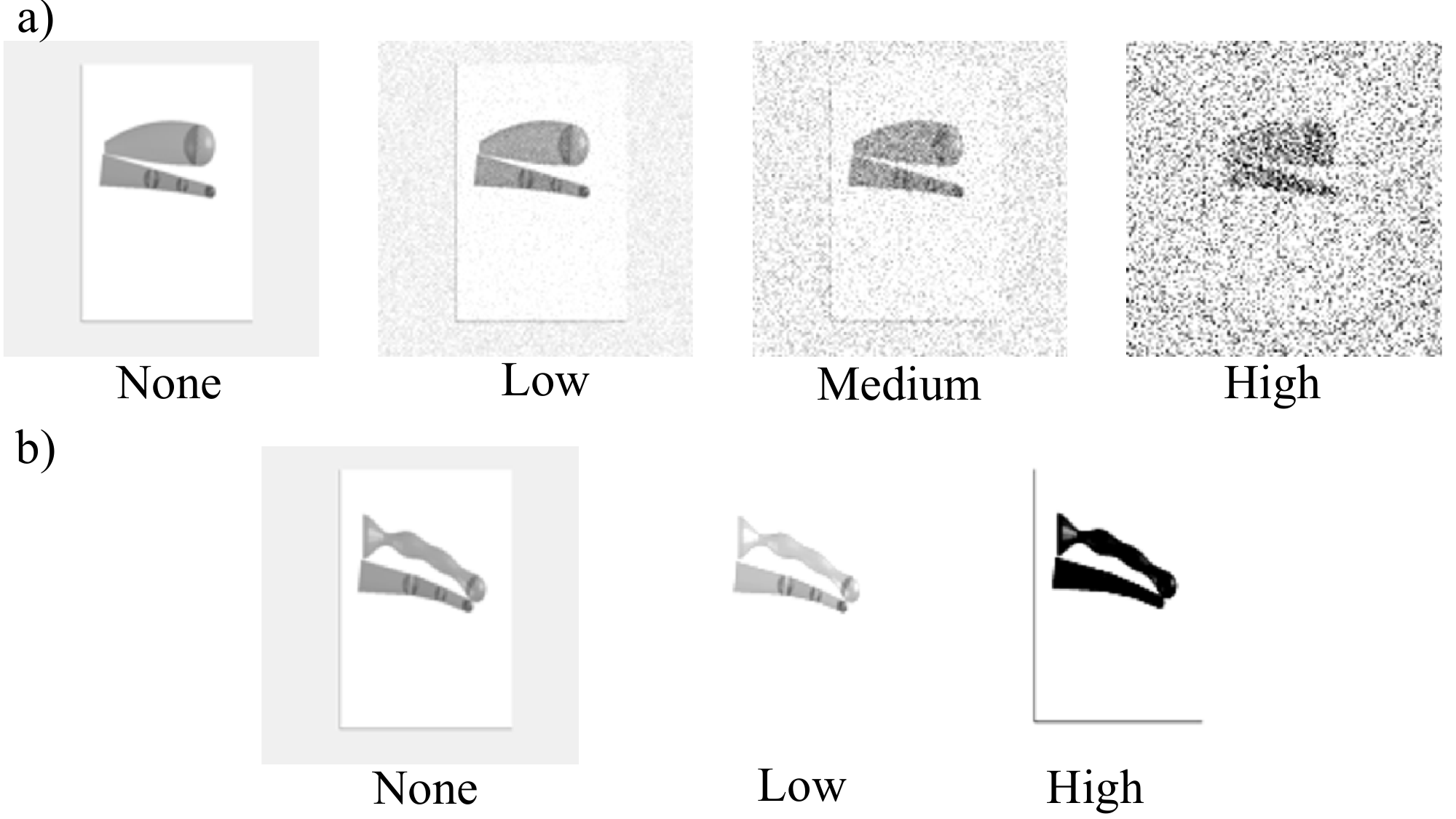}
    \caption{Adulterating the test dataset with (a) gaussian white noise with different variances (Low = 0.001, Medium = 0.01, and High= 0.1) or (b) converted to low or high contrast images.}
    \label{fig_noise_intensity_example}
\end{figure}

The generalizability of the CNN force estimator was assessed by contaminating the test set (Fig. \ref{fig_noise_intensity_example}). The MATLAB function \textbf{imnoise} was used to add gaussian white noise ($\mu$ = 0.01) with low, medium, or high variances (0.001, 0.01, and 0.1 respectively) to the frames. We also tested the performance of the CNN force estimator to varying image contrast. The pixel values were mapped from [1, 256] to [26, 200] and [174, 240] respectively for low and high contrast using the MATLAB function \textbf{imadjust}. 

\subsection{Feedback controller embedded with CNN force estimator} \label{sec_feedback_controller}
\begin{figure}[t!]
    \centering
    \includegraphics[trim = 8cm 10.5cm 5cm 0,clip,width = 0.7\textwidth]{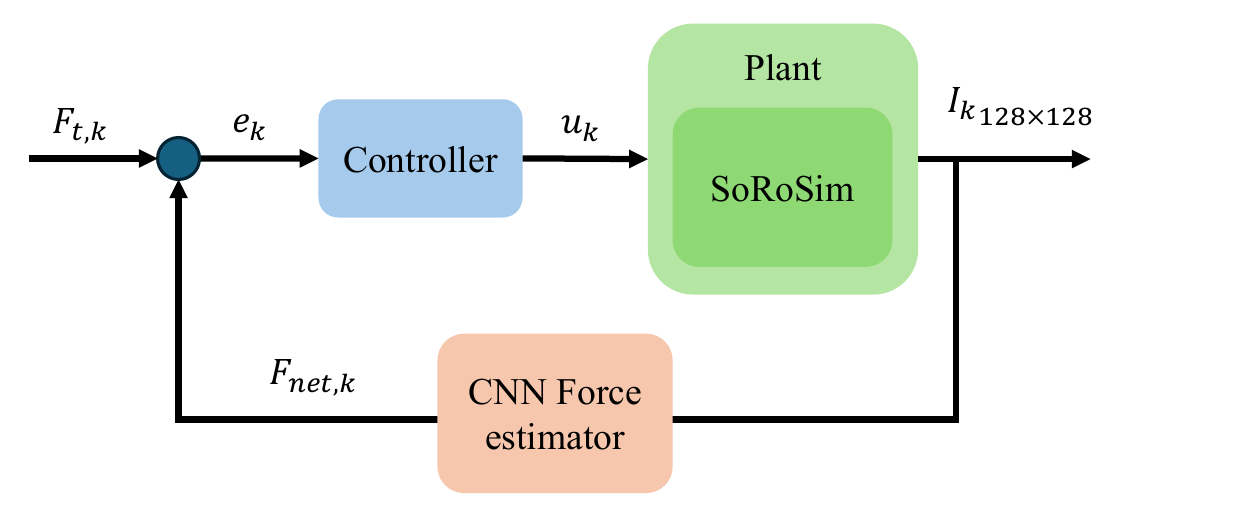}
    \caption{Feedback controller with the CNN force estimator as a force sensor. The target force ($F_t$) is tracked using a proportional controller, which estimates actuation $u$ for driving the finger-exosuit model in SoRoSim. A low-resolution grayscale snapshot of the resulting pose ($I_{128 \times 128}$) is fed to the CNN force estimator which estimates the contact force at each $C_i$ along the finger, which is summed to get the $F_{net}$. }
    \label{fig_force_controller}
\end{figure}

The CNN force estimator was integrated as a force sensor within a feedback controller as seen in Fig. \ref{fig_force_controller}. For these tests, we defined the $S(X)$ with eqn.  (\ref{eqn_R01}), with $R_0$, $R_1$, and \(L\) as 1 cm, 0.5 cm, and 9.5~cm respectively. The equations describing the controller are as follows:
\begin{eqnarray}
	 F_{net} &=& \sum_{i=1}^{7} \lVert\bm{f}_{C_i}\rVert, \; \text{and}\label{eqn_Fnet} \\
     u_k &=& u_{k-1} + K_P \cdot e_k. \label{eqn_u(t)}
\end{eqnarray}


The controlled variable was the sum of contact forces ($ F_{net}$) across all spheres on the finger (Fig. \ref{fig_use_case}). The error $e_k$ at step $k$ between the force measured by the CNN force estimator and the target force $F_t$ was implemented in a simple proportional control law to estimate the control input ($u_k$). $u_k$ was then used to actuate the exosuit. A low-resolution snapshot of the resulting pose ($\textit{I}_{128 \times 128}$) is then sent to the CNN force estimator to estimate the $ F_{net}$.

The steady-state and step responses were studied. For the steady-state test, three $F_t$ (0.25, 0.3, and 0.35 N) were defined. Here, the controller gain $K_P$ was tuned empirically to 5. For the step response, an input of 0.2 N was provided for 10 s followed by 10 s of 0 N. Here, we tested the responses for $K_P$ = 0.1, 1, and 100.  

\subsection{Analysis of Results} \label{sec_analysis_results}
We evaluate and describe the training performance of the CNN, followed by the performance on the test dataset. We show the overall performance using correlation and violin plots~\citep{Bechtold2016}, as well as a sample-by-sample comparison for randomly selected frames. The overall performance is also reported using root mean square of the differences (RMSE) between the predicted and actual force values measured at each $C_i$. Additionally, we report the RMSE normalized to the range of the target force values (RMS\%) and Pearson's correlation for each $C_i$. The RMSE, RMS\%, and CORR are then reported for the generalizability tests on contaminated images. We compare and discuss the performance of the CNN force estimator for both the steady-state and step responses of the feedback controller. All analyses were performed on MATLAB 2024 (Mathworks MA) running on a 13 Gen Intel PC with 32 RAM with 2.4 GHz. The CNN model was transformed from python to MATLAB using the function \textbf{importNetworkFromPyTorch}.

\section{Results}
\label{sec:results}
\begin{figure}[t!]
    \centering
    \includegraphics[width = 0.4\textwidth]{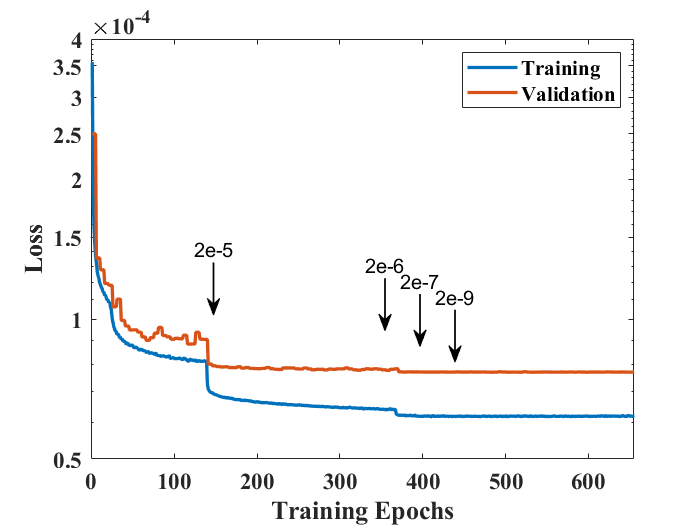}
    \caption{Evolution of the loss functions. The learning rate initialized at $2e-4$ reduces when plateauing.}
    \label{fig_losses}
\end{figure}

The progression of training and validation losses are seen in Fig. \ref{fig_losses}. The validation loss has saturated after 157 epochs with a slight improvement after 371 epochs when the learning rate was reduced to $2e-6$. The training stopped at 656 epochs due to no improvement in the validation loss. The training took about 270 minutes. 

\begin{table}[t!]
	\caption{CNN performance for $f_{C_i}$}
	\centering
	\begin{tabular}{|l|c|c|c|c|c|c|c|}
		\hline
		\textbf{Metric} & 
		$C_{1}$  & $C_{2}$ & $C_{3}$ & $C_{4}$ & $C_{5}$ & $C_{6}$ & $C_{7}$     \\ \hline
	RMSE (N)	& 	0.02	& 	0.01	& 	0.06	& 	0.07	& 	0.06	& 	0.07	& 	0.01	\\  \hline
RMS\%	& 	0.91	& 	0.64	& 	1.6	& 	2.64	& 	3.89	& 	3.85	& 	5.3	\\  \hline
CORR	& 	0.92	& 	0.99	& 	0.92	& 	0.82	& 	0.78	& 	0.51	& 	0.93	\\  \hline 
	\end{tabular}
	\label{tab_performance_base}
\end{table}

\begin{table}[t!]
	\caption{Average performance across $C_i$ for adulterated test images}
	\centering
	\begin{tabular}{|l|c|c|c|c|}
		\hline
		\textbf{Metric} & \textbf{None} &
		\textbf{Low}  & \textbf{Medium} & \textbf{High}      \\ \hline \hline
        \multicolumn{5}{c}{\textit{Gaussian Noise}} \\ \hline
	RMSE (N)	& 	0.04 ± 0.03	& 	0.05 ± 0.03	& 	0.07 ± 0.03	& 	0.09 ± 0.04	\\  \hline
RMS\%	& 	2.72 ± 1.78	& 	3.03 ± 2.03	& 	5.14 ± 3.98	& 	6.42 ± 4.65	\\  \hline
CORR	& 	0.83 ± 0.16	& 	0.82 ± 0.16	& 	0.56 ± 0.19	& 	0.11 ± 0.07	\\  \hline \hline
 \multicolumn{5}{c}{\textit{Contrast}} \\ \hline
RMSE (N)	& 	0.04 ± 0.03	& 	0.05 ± 0.03	& -- & 	0.08 ± 0.04	\\  \hline
RMS\%	& 	2.72 ± 1.78	& 	3.58 ± 2.19	& -- & 	5.06 ± 2.56	\\  \hline
CORR	& 	0.83 ± 0.16	& 	0.78 ± 0.16	& -- &	0.77 ± 0.18	\\  \hline

	\end{tabular}
	\label{tab_performance_noise_intensity}
\end{table}

\begin{figure*}[t!]
    \centering
    \includegraphics[width = \textwidth]{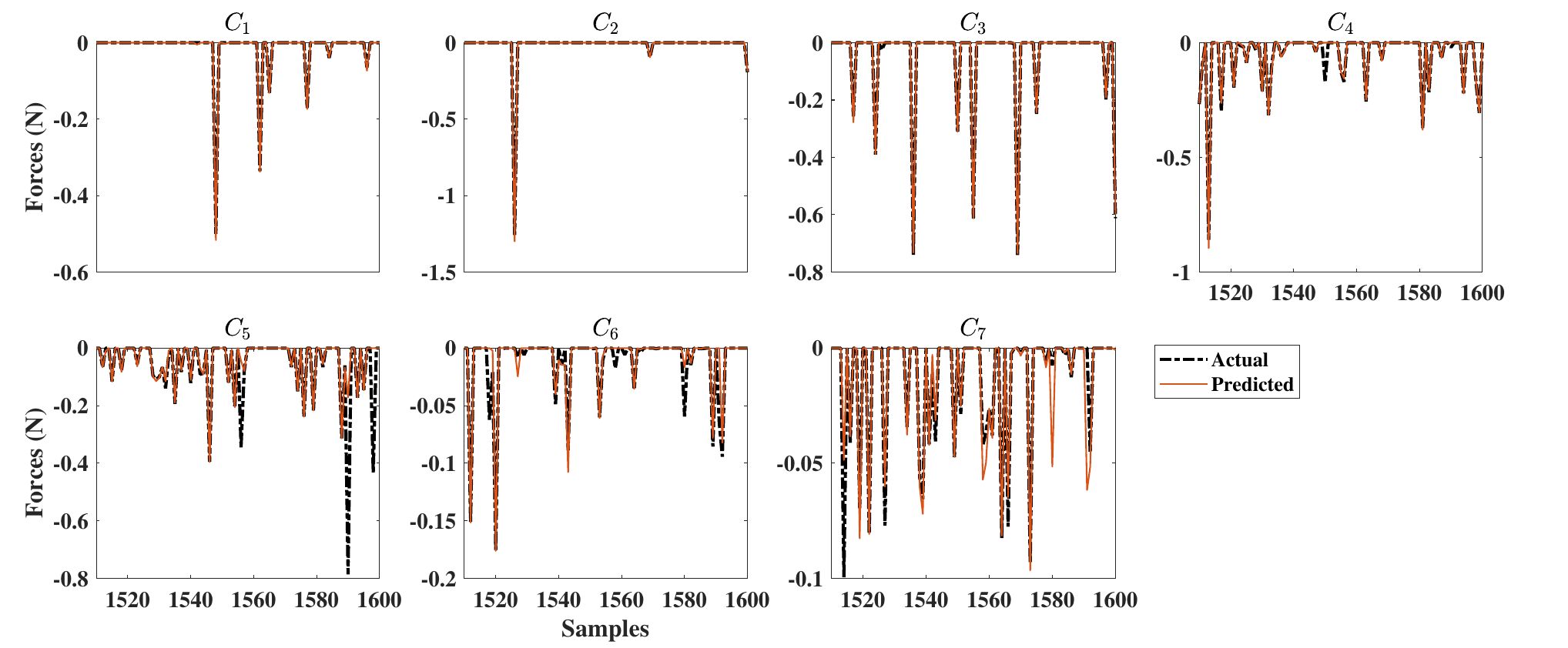}
    \caption{Performance of CNN force estimator for a random set of samples. Each panel compares the prediction of contact force with the actual values measured at $C_i$ with the SoRoSim model.}
    \label{fig_performance_zoom}
\end{figure*}

\begin{figure*}[t!]
    \centering
    \includegraphics[scale = 0.4]{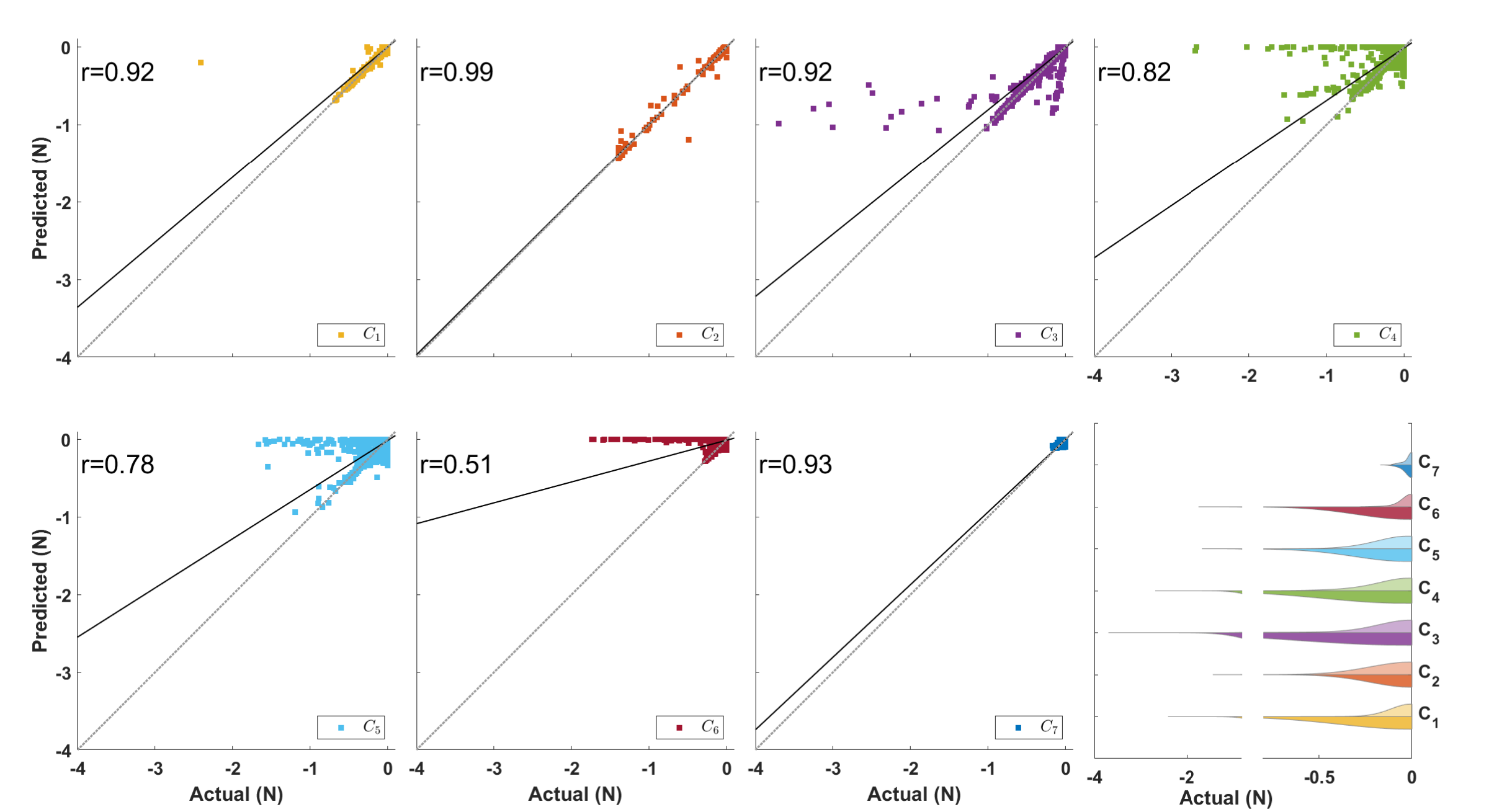}
    \caption{Overall performance of the CNN force estimator on the testing dataset. Each subplot shows the identity line and the correlation between predicted and actual forces for each $C_i$. The bottom right plot shows the raincloud distribution of the predicted (top half) versus actual (bottom half) force values.}
    \label{fig_base_ml_performance}
\end{figure*}

The performance of the CNN force estimator is shown in Fig. \ref{fig_performance_zoom},~\ref{fig_base_ml_performance}, and in Table \ref{tab_performance_base}. In Fig. \ref{fig_performance_zoom}, the performance across a few random samples from the test set is shown. We can see that the range of the contact forces measured at each $C_i$ varies. For this set of samples, the forces measured at the contact points closer to the base of the finger ($C_1$ and $C_2$) are mostly zeros. Non-zero forces are more commonly observed at $C_i$ closer to the tip of the finger-exosuit system. Overall, we see that there is good agreement between the predicted and actual forces for this set of samples. Fig. \ref{fig_base_ml_performance} demonstrates the overall performance of the CNN force estimator across each $C_i$. Across all $f_{C_i}$ in the test set, we see a moderate to strong correlation between the predicted and actual values. The violin plots show that the predicted values have a narrow distribution compared to the actual force values. The performance of the model was best for $f_{C_2}$ and lowest for the $f_{C_6}$. $C_2$ and $C_6$ are respectively, the contact spheres at the bases of the second phalange and the fingertip (Fig. \ref{fig_use_case}). The predicted and actual maximum contact forces for $C_1$ were 0.69 N and 2.41 N respectively. Similarly, for $C_2$ it was 1.43 N and 1.4 N, $C_3$ was 1.07 N and 3.7 N, $C_4$ was 0.95 N and 2.69 N, $C_5$ was 0.94 N and 1.67 N, $C_6$ was 0.28 N and 1.74 N, and $C_7$ was 0.11 N and 0.17 N respectively. The minimum value of the actual and predicted forces were 0 N for all $f_{C_i}$. 
Table \ref{tab_performance_base} shows that the errors are generally quite small across $C_i$ and are at worst 5.3\% of the actual force values. The RMS\% gets worse as $C_i$ is closer to the tip, but the trend for CORR is not similar because of the variability in the range of forces measured at the contact points. 

Table \ref{tab_performance_noise_intensity} demonstrates that the model is able to estimate the contact forces even though the contaminated images were not part of the training or validation routines. The model performs reasonably well even with medium gaussian noise (var = 0.01), and performs worse for higher than lower contrast images.

\begin{figure}[t!]
    \centering
    \includegraphics[trim = 0.2cm 0 0 0,clip,scale = 0.6]{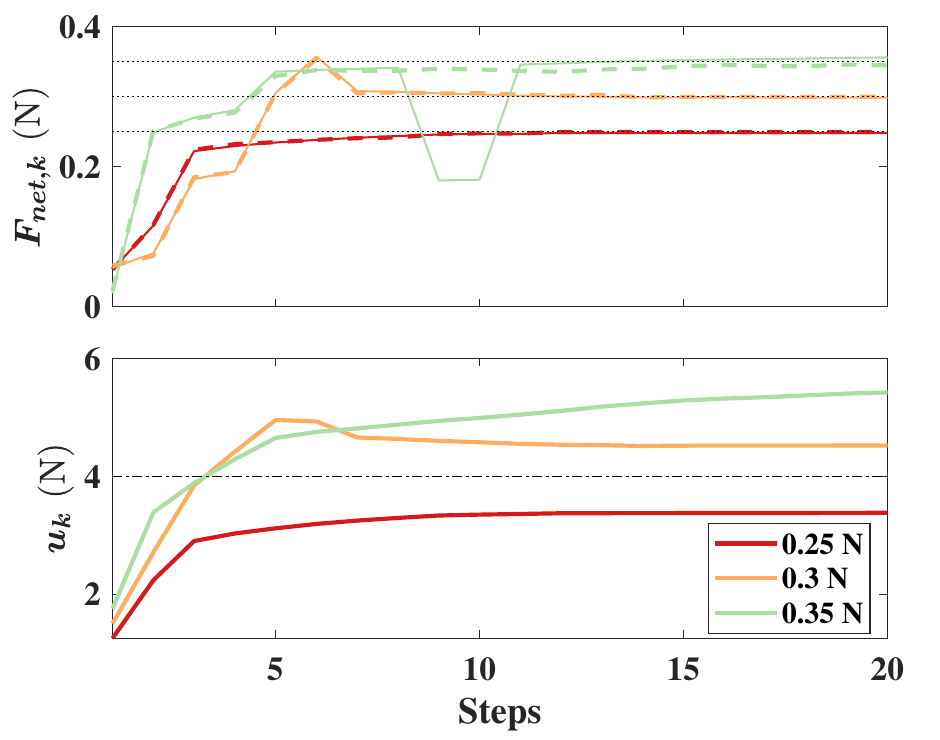}
    \caption{Steady-state responses of the feedback controller. In the top plot, the dotted lines denote $F_{net}$ measured by the CNN force estimator, whereas, the solid lines denote the reference from SoRoSim. The $u_k$ is shown in the second plot. The black dotted line denotes the maximum actuation (4 N) used for generating the training datasets.}
    \label{fig_force_controller_track}
\end{figure}


We proceed to test the CNN force estimator as a force sensor within a feedback controller (Fig. \ref{fig_force_controller}). The figure shows that for all $F_t$, the $F_{net}$ measured by the CNN force estimator overlaps with the reference SoRoSim model estimates. 
Steady state 
was reached in 7, 5, and 14 steps for 0.25, 0.3, and 0.35 N respectively. 
For $F_t$ of 0.35 N, the SoRoSim toolbox estimates steps with discrete changes in $F_{net}$, whereas, the CNN force estimations are smoother. The second plot in Fig. \ref{fig_force_controller_track} shows the $u$ used to actuate the exosuit. We see that the CNN force estimator was able to estimate the $F_{net}$ even with poses where the exosuit actuation was greater than 4 N. This is noteworthy, since the training dataset did not include actuation beyond 4 N. We found that increasing the $K_P$ reduced the steps needed to reach steady state for higher $F_t$, while worsening performance for lower $F_t$. 

\begin{figure}[t!]
    \centering
    \includegraphics[trim = 6.5cm 0 5cm 0,clip, width = 0.7\textwidth]{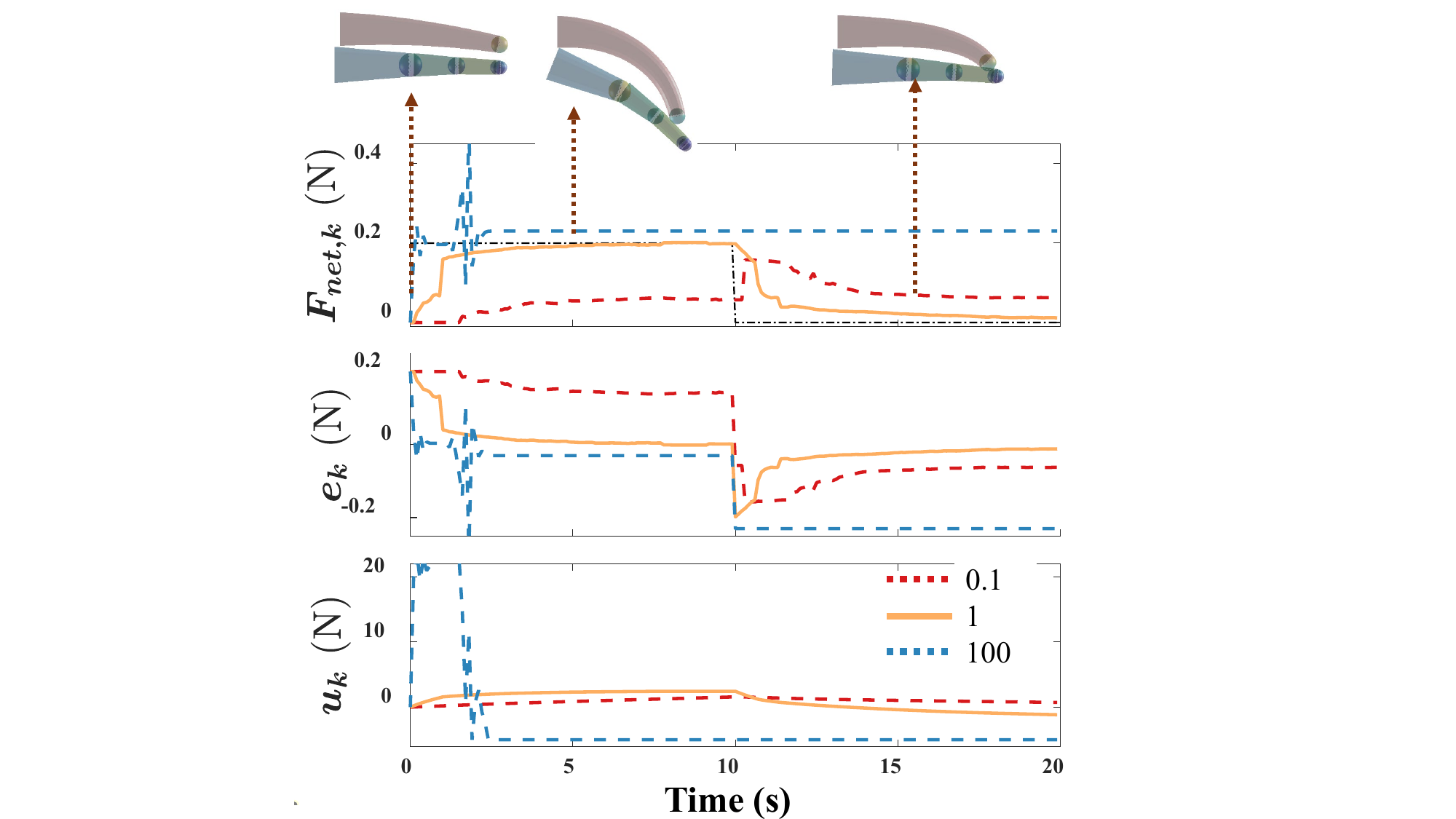}
    \caption{Step responses of the feedback controller. The first plot shows the measured $F_{net}$. The black dotted line denotes the target $F_t$. The second plot shows the error $e$, and the third plot shows the estimated $u$. The poses of the finger-exosuit system for $K_P$ = 1 at different instances (t = 0, 5, and 10 s) are shown at the top.}
    \label{fig_force_controller_step}
\end{figure}

In Fig. \ref{fig_force_controller_step}, we show the step response of the feedback controller. The target $F_{t,k}$ is seen as black dotted lines in the first plot. $K_P$ of 1 was best in tracking the step input and bringing the exosuit back to initial pose after the actuation is switched off. This is visible in the inset frames shown on top of the figure. The controller provides an $u_k$ of 2.4 N for the exosuit to exert a $F_{net}$ of 0.2 N on the finger. After the $F_t$ switches to 0 N, the controller actuates the exosuit with -1 N to bend the exosuit in the opposite direction and prevent contact with the finger. With $K_P$ = 0.1, the controller takes longer to track the step input. Here, we can see that the actuation reaches at most to 1.6 N before the step input is switched off. Finally, $K_P$ = 100 results in actuation values that overshoot the $F_t$. The finger-exosuit system ends up in a pose that it cannot recover from, as can be seen as a constant $F_{net}$ of 0.23 N after 2.6 s.

\section{Discussion}
\label{sec:discussion}
This study highlights the feasibility of using image-based learning models to understand and predict human-exosuit interactions. Here, we demonstrated a CNN force estimator that mapped snapshots of finger-exosuit pose to interaction forces between them. SoRoSim toolbox \citep{Mathew2024} was used for simulating a comprehensive dataset for training the CNN. To ensure generalizability, different formulations of the exosuit surface were utilized. The findings show that the methodology has potential for translation to real-world scenarios.  

Fig. \ref{fig_base_ml_performance} and Table \ref{tab_performance_base} show that the CNN force estimator was able to reproduce the interaction forces well. The RMS\% (at most 5.3\%) shows that the errors are small compared to the overall distribution. Generally, we see in Fig. \ref{fig_base_ml_performance} that the model predicts forces to be closer to zero compared to the distribution of the actual force values. The CNN was not good with mapping outlier force values either due to the lack of examples in the dataset or because of erroneous estimation by the simulation toolbox itself (such instances are observed at high force values).  

The CNN-based force estimator was able to abstract sufficient information from the images to be generalizable (Table \ref{tab_performance_noise_intensity}). Adding low or medium gaussian noise, or reducing image contrasts did not affect the performance of the model. However, deterioration of image quality (Fig. \ref{fig_noise_intensity_example}), and thereby features, reduced the performance. This was seen for very noisy or high contrast images.  

We integrated the CNN model as a force sensor within a feedback controller. Irrespective of the target being tracked in Fig. \ref{fig_force_controller_track}, the CNN performance was similar to the reference SoRoSim values. Strikingly, the CNN force estimator was able to estimate the interaction forces when the exosuit was actuated beyond 4 N, even though these values were not included in the training dataset. This suggests that the CNN was able to generalize a relationship between pose and net interaction force from the low-resolution grayscale images, and extrapolate to poses not included in the training dataset. Moreover, for higher actuation values, we observe benefits over SoRoSim estimation. For certain actuation values, the simulation platform measures a drop in interaction force erroneously, whereas, the CNN force estimator measures a continuously smooth change in the net force estimates. 

Modelling the interface dynamics between assistive devices and human is necessary for designing efficient systems \citep{Bardi2022}. Studies show that more than half of the assistive power can be lost in the soft interface between the exosuit and user \citep{Yandell2017}. Increasing the stiffness of the interface further reduces the power available for joint assistance \citep{Barrutia2024}. However, modelling the interaction forces is not trivial in real-world scenarios \citep{Xiloyannis2018, Andrikopoulos2015}. Although, there have been studies using computer vision to predict interaction forces between humans (or soft-robots) and objects \citep{ Pham2018, Kennedy2005, Huang2024}, they do not consider assistive devices. 
In this study, we demonstrate the potential of using images to predict interaction forces through a learning-based approach. Notably, our model exhibits strong generalization capabilities, successfully capturing interaction forces even when image quality is degraded. This suggests its applicability in real-world scenarios where camera data may be noisy or of low resolution. Furthermore, we show that the model can generalize to higher actuation forces that were not represented in the training data, indicating its robustness across a broader range of operating conditions. Finally, when integrated into a feedback control system, the model functions as a surrogate force sensor, offering a practical alternative to physical force sensing.




There are a few limitations to this study. The study offers insights only based on simulation trials, and translation to real-world scenarios and validating the findings is necessary to continue translation to actual practice. The current study modelled the finger-exosuit interface using a simple contact model with normal forces. Moreover, the system's static equilibrium was solved for each pose with increasing actuation for the exosuit. Tangential friction and velocity-dependent interactions could be included to improve the interaction force estimations \citep{Flores2022, Menager2025}. 
Additionally, we only simulated variations in the exosuit shape, with constant dimensions of the finger joint. Inclusion of different finger dimensions could improve the variability of the dataset.
Future work will focus on extending the approach to multi-joint exosuits and incorporating real-world data to validate the model beyond simulated environments. Additionally, exploring reinforcement learning techniques to refine exosuit control strategies based on the learned interaction forces presents an exciting avenue for further research \citep{Caggiano2022}.

The study presents a robust, data-driven framework for modeling human-exosuit interfaces using simulation and deep learning. Its generalizability shows promise for real-world application, including low-resolution imaging scenarios, while sufficient training data may reduce reliance on precise interface modeling or force sensors.
\section*{Acknowledgment}
This work was partly supported by the Netherlands Sectorplan Techniek 2 and the Royal Society research grants RGS/R1/231472 and IES/R2/242059. The authors thank Yunqi Huang for fabricating the exosuit, and Arvid Keemink for the training infrastructure.




%
\bibliographystyle{unsrtnat} 
\bibliography{main_paper}

%




\end{document}